\documentclass[letterpaper]{article} 
\usepackage[draft]{aaai2026}  
\usepackage{times}  
\usepackage{helvet}  
\usepackage{courier}  
\usepackage[hyphens]{url}  
\usepackage{graphicx} 
\usepackage{natbib}  
\usepackage{caption} 
\usepackage{algorithm}
\usepackage{algorithmic}

\usepackage{amsmath,amsfonts}
\usepackage{array}
\usepackage{subfig}
\usepackage{textcomp}
\usepackage{url}
\usepackage{verbatim}
\usepackage{graphicx}
\usepackage{cite}
\usepackage{xcolor}
\usepackage{soul}
\usepackage{amssymb}
\usepackage{enumitem}

\usepackage{arydshln}
\usepackage{fancyhdr}
\usepackage{varwidth}
\usepackage{arydshln}
\usepackage{multirow}
\usepackage{mathtools}
\usepackage{algorithmic}
\usepackage{algorithm}

\usepackage{datetime2}
\usepackage{tabularray}

\usepackage{newfloat}
\usepackage{listings}
\usepackage{pifont}

\usepackage{kotex}
\usepackage{soul}

\usepackage{tikz}
\newcommand{\circled}[1]{\tikz[baseline=(char.base)]{
  \node[shape=circle,draw,inner sep=0.5pt] (char) {\small #1};}}

\newcommand{\sys}{STAS}
\newcommand{\syss}{STAS$^{\epsilon}$}

\newcommand{\remove}[1]{}

\usepackage{newfloat}
\usepackage{listings}
\DeclareCaptionStyle{ruled}{labelfont=normalfont,labelsep=colon,strut=off} 
\floatstyle{ruled}
\newfloat{listing}{tb}{lst}{}
\floatname{listing}{Listing}
\title{STAS: Spatio-Temporal Adaptive Computation Time for Spiking Transformers}
\author{
    Donghwa Kang\textsuperscript{\rm 1}, Doohyun Kim\textsuperscript{\rm 1}, Sang-Ki Ko\textsuperscript{\rm 2}, Jinkyu Lee\textsuperscript{\rm 3,}\thanks{Jinkyu Lee initiated this work while affiliated with Sungkyunkwan University.}, Brent ByungHoon Kang\textsuperscript{\rm 1}, \\ Hyeongboo Baek\textsuperscript{\rm 2}
}
\affiliations{
    \textsuperscript{\rm 1}Korea Advanced Institute of Science and Technology (KAIST) \\
    \textsuperscript{\rm 2}University of Seoul \\
    \textsuperscript{\rm 3}Yonsei University

}

\usepackage{bibentry}

\begin{document}

\maketitle

\begin{abstract}
Spiking neural networks (SNNs) offer energy efficiency over artificial neural networks (ANNs) but suffer from high latency and computational overhead due to their multi-timestep operational nature. 
While various dynamic computation methods have been developed to mitigate this by targeting spatial, temporal, or architecture-specific redundancies, they remain fragmented. 
While the principles of adaptive computation time (ACT) offer a robust foundation for a unified approach, its application to SNN-based vision Transformers (ViTs) is hindered by two core issues: the violation of its temporal similarity prerequisite and a static architecture fundamentally unsuited for its principles.
To address these challenges, we propose \sys{} (\textbf{S}patio-\textbf{T}emporal \textbf{A}daptive computation time for \textbf{S}piking transformers), a framework that co-designs the static architecture and dynamic computation policy. 
STAS introduces an integrated spike patch splitting (I-SPS) module to establish temporal stability by creating a unified input representation, thereby solving the architectural problem of temporal dissimilarity. 
This stability, in turn, allows our adaptive spiking self-attention (A-SSA) module to perform two-dimensional token pruning across both spatial and temporal axes. 
Implemented on spiking Transformer architectures and validated on CIFAR-10, CIFAR-100, and ImageNet, STAS reduces energy consumption by up to 45.9\%, 43.8\%, and 30.1\%, respectively, while simultaneously improving accuracy over SOTA models. 
\end{abstract}

\section{Introduction}
\label{sec:intro}

\begin{figure}[t!]
    \centering
    \includegraphics[width=1\linewidth]{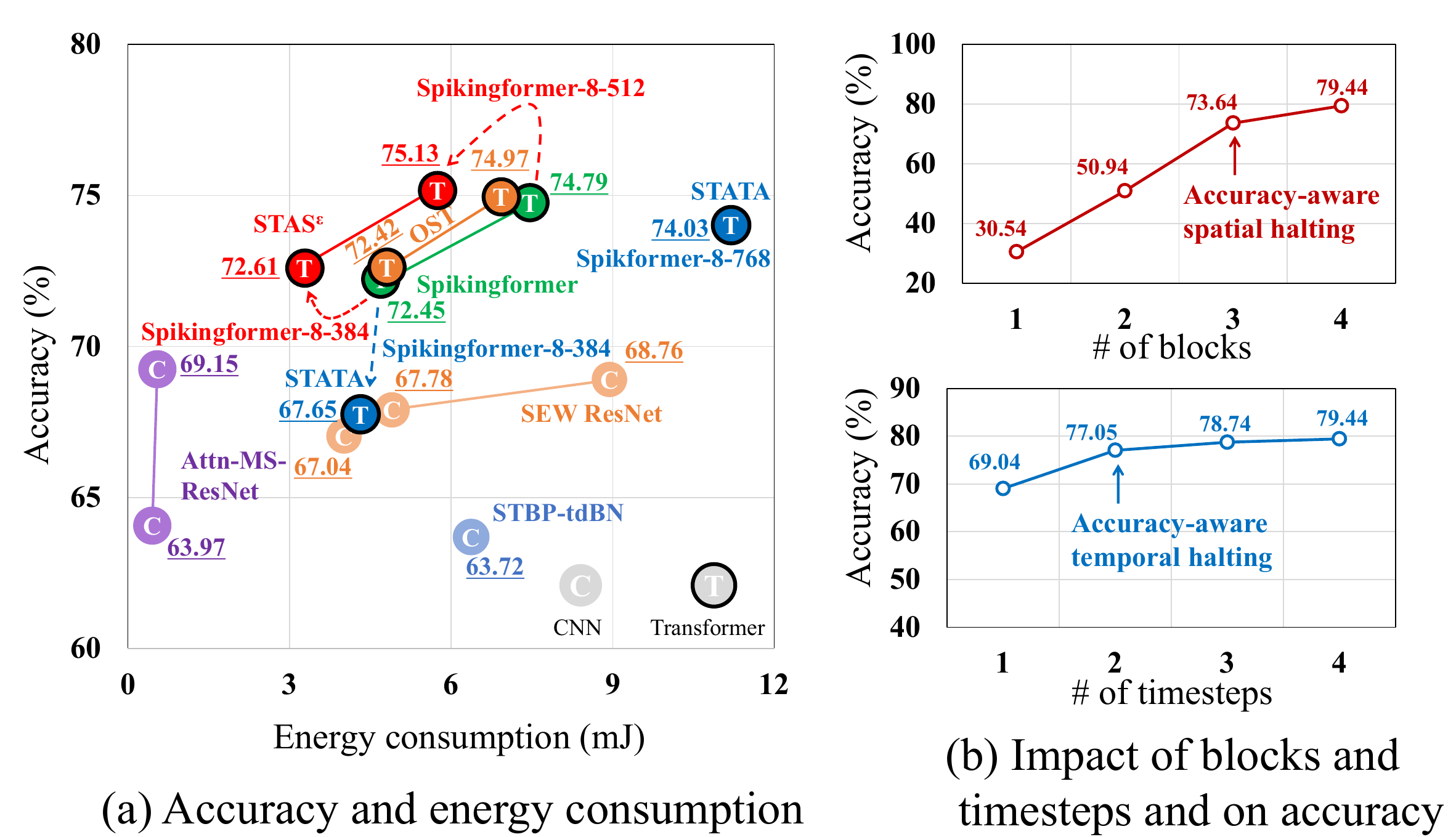}
    \caption{Accuracy comparison of adaptive computation methods for Spikingformer on ImageNet. (a) Accuracy-energy trade-off, where each method may represent multiple model sizes (corresponding to Table~\ref{tab:main_exp}). (b) Accuracy saturation motivating accuracy-aware halting spatially after the 3rd block (top) and temporally after the 2nd timestep (bottom).}
    \label{fig:intro}
    \vspace{-0.2cm}
\end{figure}

Spiking neural networks (SNNs) offer energy efficiency over artificial neural networks (ANNs) but suffer from high latency and computational overhead due to their multi-timestep operational nature. 
To overcome this limitation, state-of-the-art (SOTA) study to improve SNN efficiency has largely followed two paths: \texttt{\textbf{(S)}} static architectural (i.e., model) enhancements, which redesign core components (e.g., Spikformer~\cite{ZZH22}, Spikingformer~\cite{ZYZ23}), and \texttt{\textbf{(D)}} dynamic computation methods, which optimize for computational redundancy (e.g., OST~\cite{SSX24}, STATA~\cite{ZWY24}), with their performances illustrated in Fig.~\ref{fig:intro}(a). 
The motivation for such dynamic methods is illustrated in Fig.~\ref{fig:intro}(b), which shows that accuracy on a given task frequently saturates long before the final computational block (spatially) or timestep (temporally), presenting a clear opportunity for input-dependent, accuracy-aware halting. 

The exploration of dynamic computation has itself fragmented into distinct approaches. 
One line of research, primarily in conventional ANNs, has refined \texttt{\textbf{(D1)}} spatial halting, the concept of pruning computation across layers, which is largely architecture-agnostic (e.g., SACT~\cite{FCZ17}).
In parallel, SNN-specific works have focused on \texttt{\textbf{(D2)}} temporal adaptivity, dynamically adjusting the number of timesteps based on input difficulty (e.g., DT-SNN~\cite{LMG23}). 
A third, more sophisticated approach is \texttt{\textbf{(D3)}} architecture-aware halting, which leverages the unique computational units of a model, such as the token-level processing in Transformers (e.g., A-ViT~\cite{YVA22}). 
This separation suggests that a paradigm capable of holistically integrating these spatial, temporal, and architecture-aware strategies \texttt{\textbf{(D1-D3)}} could yield substantial efficiency gains. 
The primary challenge, therefore, lies in developing a unified framework that can navigate the complex interplay between these adaptive approaches.

While we argue that the principles of adaptive computation time (ACT) offer a robust foundation for a unified dynamic framework within SNN-based vision Transformers (ViTs), a purely dynamic approach faces a two-fold challenge. 
First, the principles of ACT, developed for single-path networks like RNNs~\cite{GRA16} and ViTs, have not been validated for the discrete temporal domain of SNNs. 
Second, even if this extension were possible, its efficacy would be fundamentally constrained by the underlying static architecture. Specifically, the principles of ACT, as demonstrated in conventional ViTs, depend on high input similarity via a residual network to ensure stable representation refinement (Fig. 2(b), 2(d)). 
While SNN-based ViTs possess the necessary spatial similarity for block-level halting (Fig. 2(e), left), their inherent architectural design leads to critically low temporal similarity due to varying spike inputs at each timestep (Fig. 2(e), right). 
This architectural flaw obstructs any temporal or spatio-temporal halting, revealing a critical interdependence: a novel static architecture is a prerequisite for an effective dynamic computation framework, necessitating a new, integrated paradigm \texttt{\textbf{(S with D1-D3)}}.

In this paper, we propose STAS (\textbf{S}patio-\textbf{T}emporal \textbf{A}daptive computation time for \textbf{S}piking Transformers), a novel framework that resolves this challenge by co-designing the static architecture and the dynamic computation method. 
STAS first addresses the architectural bottleneck with a new architectural module, the integrated spike patch splitting (I-SPS), which provides the static solution \texttt{\textbf{S}} by creating a temporally unified and consistent representation. 
This engineered temporal stability, in turn, enables our novel dynamic halting mechanism, the adaptive spiking self-attention (A-SSA), to act as the unified framework for \texttt{\textbf{D1-D3}} by performing concurrent, accuracy-aware token halting across both spatial and temporal axes.

We implemented \sys{} on strong architecture with STBP-based direct training, including Spikformer and Spikingformer, and validated its performance on the CIFAR-10, CIFAR-100, and ImageNet classification datasets. 
When applied to spiking Transformer architectures on the three datasets, STAS reduces energy consumption by up to 45.9\%, 
43.8\%, and 30.1\%, respectively, while simultaneously improving top-1 accuracy. 

Our contribution can be summarized as follows:
\begin{itemize}
    \item We diagnose the fundamental barrier to a unified adaptive framework in SNN-based ViTs through a spatio-temporal similarity analysis, revealing that their architectural design inherently obstructs temporal halting.
    \item We propose I-SPS that re-engineers the SNN input stage to establish the temporal similarity required for effective temporal adaptation.
    \item Building upon the stability provided by I-SPS, we introduce A-SSA, a unified mechanism that performs concurrent spatial and temporal token halting.
    \item We demonstrate the effectiveness of STAS through extensive experiments on CIFAR-10, CIFAR-100, and ImageNet, achieving up to 45.9\%, 43.8\%, and 30.1\%, respectively, for SOTA architectures while improving accuracy.
\end{itemize}

\section{Related Work}
\label{sec:related_work}

Methods like DT-SNN dynamically adjust the timesteps of an SNN during inference based on accuracy needs, using entropy and confidence metrics to halt computation early for simpler inputs. SEENN~\cite{LMG23,LGK24} employs reinforcement learning to optimize timesteps for each image, allowing for fine-grained per-instance optimization, while TET~\cite{DLZ22} introduces a loss function to address gradient loss in spiking neurons, achieving higher accuracy with fewer timesteps. However, the decision-making overhead of these temporal methods can outweigh the benefits in low-timestep regimes, making them less suitable for deeper, more efficient models. In a different approach, MST~\cite{WFC23} proposes an ANN-to-SNN conversion method for SNN-based ViTs, using token masking within model blocks to reduce energy consumption. Despite its effectiveness, this reliance on ANN-to-SNN conversion means MST still requires hundreds of timesteps for inference.

The principles of ACT~\cite{GRA16} were first introduced to dynamically allocate inference steps for RNN models based on input difficulty. This concept was extended by SACT~\cite{FCZ17} for ResNet architectures and A-ViT~\cite{YVA22}, which dynamically adjusts computation in Transformers by halting individual tokens at different layers. However, these studies are based on ANNs, and their formulations are fundamentally incompatible with the discrete, multi-timestep nature of SNNs, as they typically perform a single inference pass. While LFACT~\cite{ZEK21} expands ACT for repeated inferences across sequences, it remains limited to RNNs. In contrast, STAS is explicitly designed to address the unique two-dimensional challenge of SNN-based ViTs, simultaneously considering adaptivity across both spatial blocks and discrete timesteps.

\begin{figure}[t!]
    \centering
    \includegraphics[width=1\linewidth]{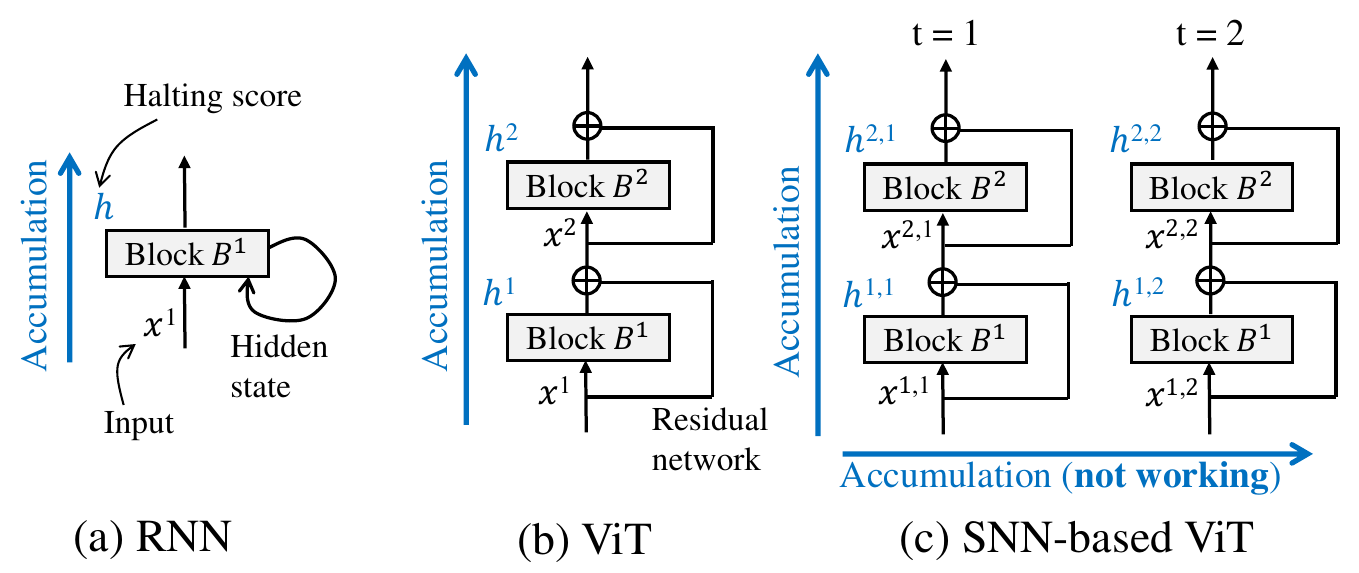}    
    \includegraphics[width=1\linewidth]{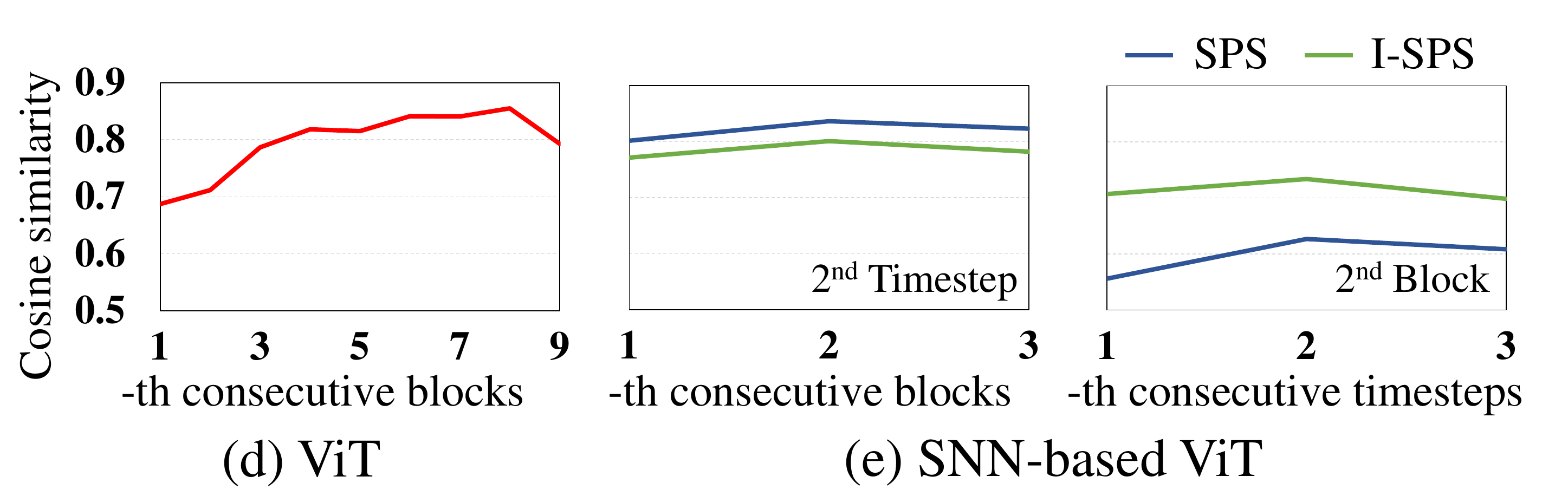}
    \caption{Model architecture and halting-score accumulation paths when Adaptive Computation Time (ACT) is applied: (a) RNN, (b) ViT, and (c) SNN-based ViT. Cosine similarity of tokens between consecutive blocks for (d) ViT and (e) SNN-based ViT (Spikingformer) on CIFAR-100.}
    \label{fig:combined_motivation}
    \vspace{-0.2cm}
\end{figure}

\begin{figure}[t!]
    \centering
    \includegraphics[width=1\linewidth]{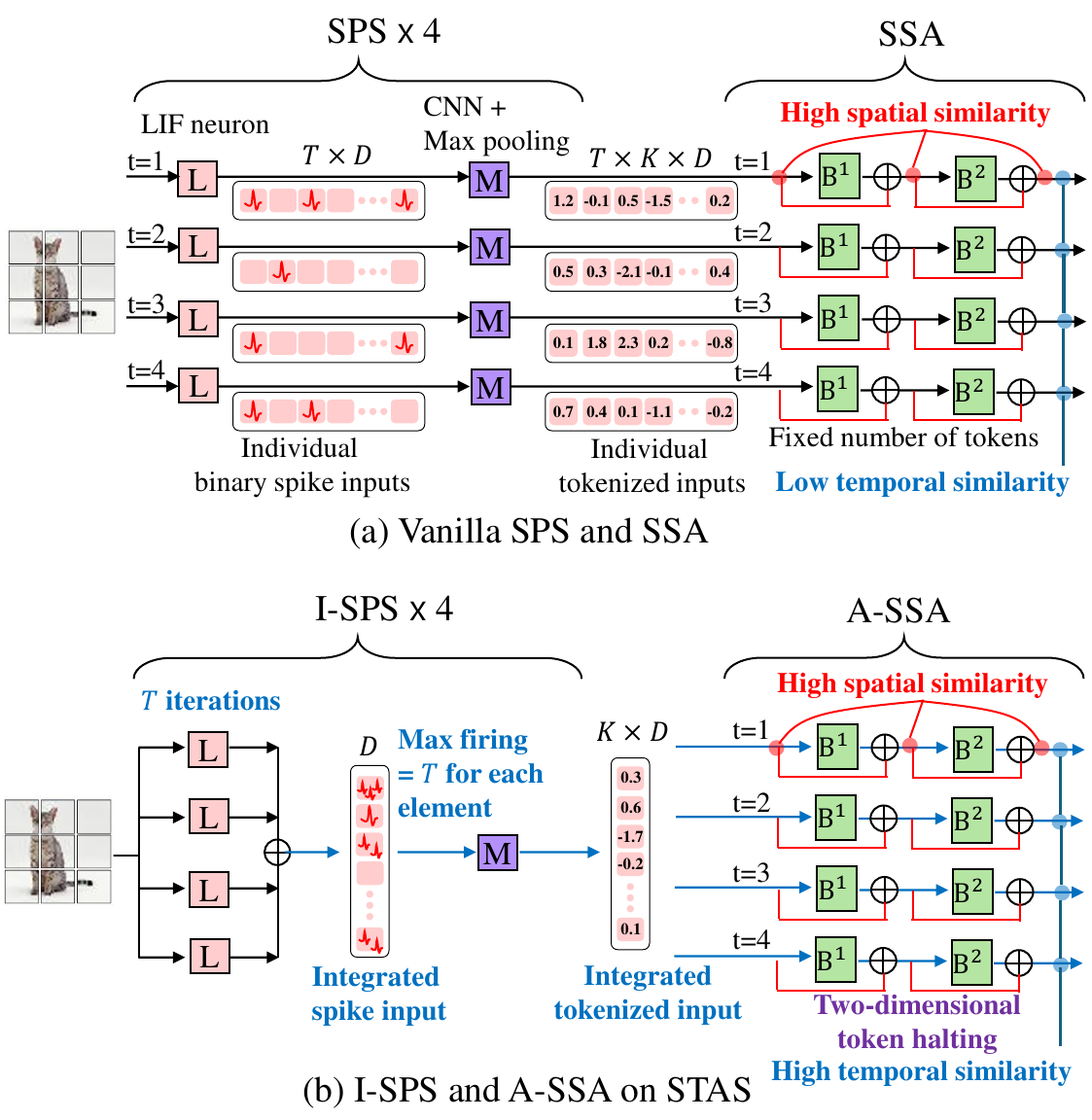}
    \caption{Architectural comparison of (a) a conventional SNN-based ViT using vanilla SPS and SSA, and (b) our STAS framework featuring I-SPS and A-SSA. STAS utilizes I-SPS to create a single, unified tokenized input from multiple timesteps, which establishes the high temporal similarity necessary for the two-dimensional token halting performed by A-SSA.}
    \label{fig:SPS}
    \vspace{-0.2cm}
\end{figure}

\section{Method}
\label{sec:method}

\subsection{I-SPS: Integrated Spike Patch Splitting}
\label{subsec:i_sps}

ACT enables neural networks to dynamically adjust their computational depth per input, learning to halt processing to improve efficiency. 
The mechanism is predicated on the principle of halting computation once the network's internal representations stabilize. 
This concept was originally proposed for RNNs, where an encoder block $\mathcal{B}^1$ iteratively refines its state from the same input $x^1$, and a sigmoidal halting unit determines when to cease processing (Fig.~\ref{fig:combined_motivation}(a)). 
This architectural paradigm extends naturally to ViTs, which can be viewed as an ``unrolled iterative estimation" process. 
Their structure, featuring multiple identical encoder blocks (property (i)) with residual connections (Fig.~\ref{fig:combined_motivation}(b)), ensures high input similarity between consecutive blocks (property (ii), Fig.~\ref{fig:combined_motivation}(d)).
This representational stability is a prerequisite for ACT, enabling effective spatial halting in ViTs by allowing each block $\mathcal{B}^i$ to accumulate a corresponding halting score $h^i$~\cite{YVA22}.

However, applying ACT to SNN-based ViTs introduces a dual-dimensional challenge, as the conditions for effective halting must be met across both spatial (inter-block) and temporal (inter-timestep) axes (Fig.~\ref{fig:combined_motivation}(c)). \textbf{Spatially}, SNN-based ViTs are analogous to their standard counterparts; they satisfy property (i) via residual connections and, consequently, maintain high block-to-block similarity (property (ii)), making them suitable for spatial ACT (left subfigure of Fig.~\ref{fig:combined_motivation}(e)). \textbf{Temporally}, the challenge is more profound. While property (i) is satisfied because membrane potentials are shared across timesteps within the same block, SNNs inherently violate property (ii). Each timestep receives a different input spike vector, leading to low cosine similarity between consecutive temporal inputs, as shown by the blue curve in the right subfigure of Fig.~\ref{fig:combined_motivation}(e).

To address the low temporal similarity in SNN-based ViTs that impedes ACT, we introduce the I-SPS module. 
Unlike vanilla SPS, I-SPS integrates multi-timestep spike signals into a single, unified representation at the initial stage, which is then reused for all subsequent computations (Fig.~\ref{fig:SPS}(b)). 
This positions our method as a type of `one-step' approach\footnote{This is termed a `one-step' approach because the computationally expensive CNN operation is reduced to a single pass, while the low-latency LIF neuron operations still iterate for $T$ timesteps.}, an emerging concept in SOTA SNN studies where expensive operations are reduced to a single pass in distinct ways for varied goals, such as latency reduction (e.g., OST) or simplified adversarial attacks (e.g., RGA~\cite{BDH23}). 
The viability of such methods, which sacrifice precise temporal information, is rooted in mitigating challenges in direct SNN training; a shortened temporal backpropagation path reduces the impact of both vanishing gradients and error accumulation from surrogate functions. 
This improved gradient flow appears to offset the information loss from temporal compression. STAS operationalizes this principle via the I-SPS module, creating the high temporal similarity (Fig.~\ref{fig:combined_motivation}(e)) that is the prerequisite for our A-SSA module to perform dynamic, two-dimensional token halting.

\begin{table}[]
\small
\begin{center}
\caption{Effectiveness of I-SPS for A-SSA on Spikformer-4-384 and Spikingformer-4-384 with CIFAR-100.}
\label{tab:i_sps}
\begin{tabular}{c|cc|cc}
\hline
   Architecture & I-SPS       & A-SSA         & Avg. tokens             & Acc (\%)                   \\ \hline
\multirow{3}{*}{ {\small Spikformer}} 
      & \ding{55}   & \ding{55}       & $\times$1               & 77.3                       \\
      & \ding{55}   & \ding{51}       & $\times$0.63            & 77.3 $(-)$        \\
      & \ding{51}   & \ding{51}       & \textbf{$\times$0.46}            & \textbf{78.1 $(\uparrow)$}   \\ \hline
\multirow{3}{*}{ {\small Spikingformer}} 
      & \ding{55}   & \ding{55}       & $\times$1               & 79.4              \\
      & \ding{55}   & \ding{51}       & $\times$0.95            & 77.4 $(\downarrow)$        \\
      & \ding{51}   & \ding{51}       & \textbf{$\times$0.70}   & \textbf{79.9 $(\uparrow)$}        \\ \hline
\end{tabular}
\end{center}
\end{table}

\begin{figure*}[t]
    \centering
    \includegraphics[width=1\linewidth]{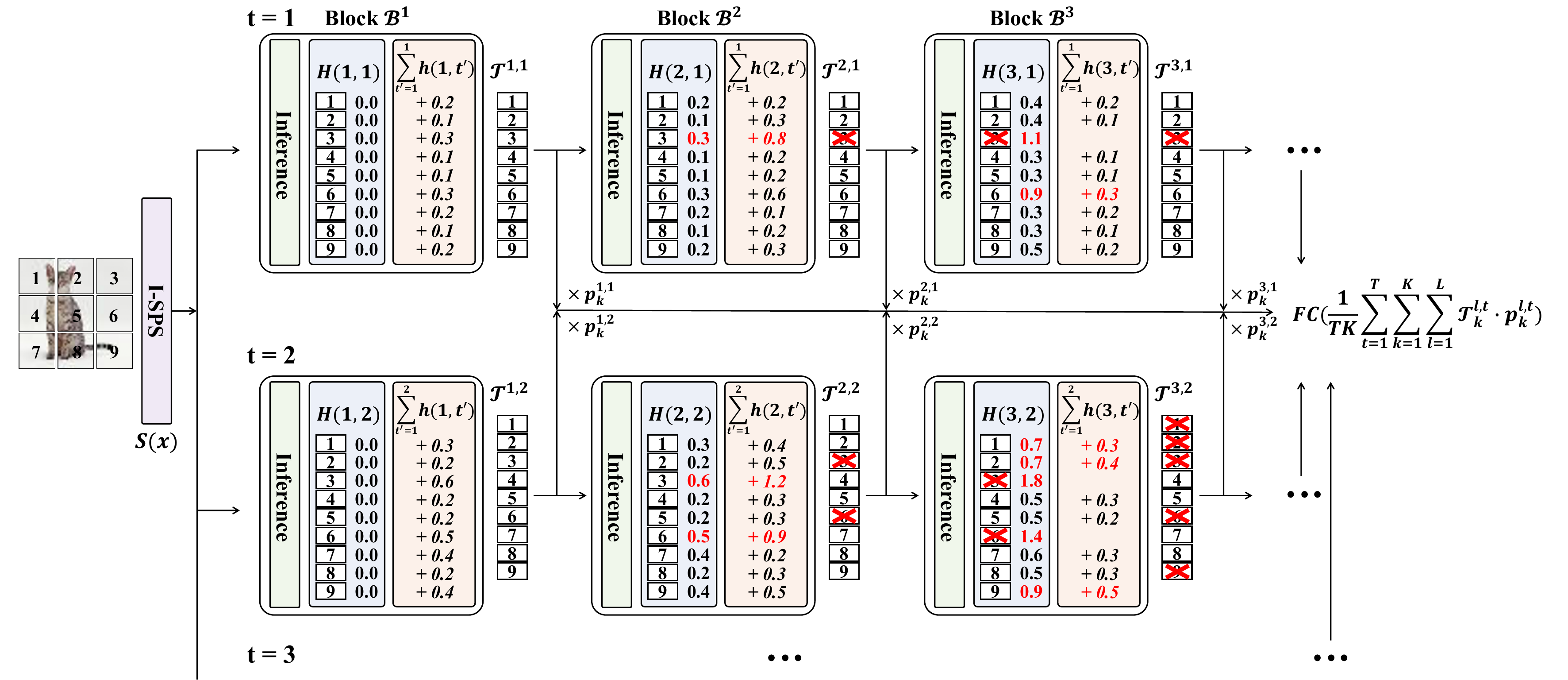}
    \caption{Token-level halting example of \sys{}: At the first timestep $t=1$, the input $x$ passes through the I-SPS, generating a token set $\mathcal{T}^{l, t}$. In the first block $\mathcal{B}^1$, for nine tokens, the halting scores $h_k^{1,1}$ are added through inference. In subsequent blocks, tokens with accumulated halting scores $H(l,t)$ of one or greater are masked.
    From the second timestep onwards, the same operations are repeated on the same input $x$. The halting score accumulation follows Eq.~\eqref{eq:M_score}. The vector values of masked tokens are set to zero, and no further halting score is accumulated for the tokens.}
    \label{fig:main_example}
    \vspace{-0.2cm}
\end{figure*}

\paragraph{Empirical validation.}
Table~\ref{tab:i_sps} validates the synergistic relationship between our static architectural module (I-SPS) and dynamic halting mechanism (A-SSA), which is detailed in Sec.~\ref{subsec:ACT}. 
Applying A-SSA alone is ineffective, yielding only a limited token reduction on both Spikeformer and Spikingformer ($\times$0.63 and $\times$0.95, respectively). 
However, when combined with I-SPS—which establishes the necessary temporal similarity—the synergy drastically reduces token usage to $\times$0.46 on Spikformer and $\times$0.70 on Spikingformer, while maintaining or even slightly improving accuracy. 
These results empirically demonstrate that I-SPS is a critical prerequisite for A-SSA to perform efficient and accuracy-aware spatio-temporal halting.

\subsection{A-SSA: Adaptive Spiking Self-Attention}
\label{subsec:ACT}

We formulate the SNN-based ViT as follows~\cite{ZYZ23}:
\begin{equation}
    f_{T}(x) = FC(\frac{1}{T} \sum^{T}_{t=1} \mathcal{B}^{L} \circ \mathcal{B}^{L-1} \circ \cdot \cdot \cdot \circ \mathcal{B}^{1} \circ \mathcal{S}(x)),
    \label{eq:gradient}
\end{equation}
where $x \in \mathbb{R}^{T \times C \times H \times W}$ is the input of which $T$, $C$, $H$, and $W$ denote the timesteps, channels, height, and width. 

The function $\mathcal{S}(\cdot)$ represents the spike patch splitting (SPS) module, which divides the input image into multiple tokens. 
The function $\mathcal{B}(\cdot)$ denotes a single encoder block, consisting of spike self-attention (SSA) and a multi-layer perceptron (MLP), with a total of $L$ blocks in the model.
The function $FC(\cdot)$ represents a fully-connected layer. 
Finally, the tokens passing through all blocks are averaged and input to $FC(\cdot)$.

After passing through $\mathcal{S}(x)$ at a timestep $t$, the input image $x$ is divided into a set of tokens denoted by $\mathcal{T}^{0,t}$. 
Let $\mathcal{T}^{l, t}$ represent the set of tokens 
in the $l$-th (for $l > 0$) block at the $t$-th timestep, which is expressed as follows: 
\begin{equation}
    \mathcal{T}^{l, t} = \mathcal{B}^{l}(\mathcal{T}^{l-1, t}).
    \label{eq:tokenes}
\end{equation}

The halting score $h^{l, t}$ of the tokens at the $t$-th timestep in the $l$-th block can be defined as follows:
\begin{equation}
    h^{l, t}_{k} = \sigma(\alpha \times \mathcal{T}^{l, t}_{k,1} + \beta),
    \label{eq:halting}
\end{equation}
where $\sigma(\cdot)$ denotes the logistic sigmoid function, and $\alpha$ and $\beta$ are scaling factors.

Let $\mathcal{T}^{l,t}_{k}$ represent the embedding vector of the $k$-th token, and $\mathcal{T}^{l,t}_{k,1}$ denote the first element of this vector.
The sigmoid function ensures that $0 \leq h^{l, t}_{k} \leq 1$. 
\sys{} calculates $h^{l, t}_{k}$ using the first element of the embedding vector of the token, and the first node of MLP in each block learns the halting score.

\sys{} accumulates halting scores across blocks within a single timestep and continues to accumulate scores from previous timesteps and blocks over multiple timesteps, as a two-dimensional halting policy.
\sys{} defines the halting module $H_{k}(L', T')$ at the $T'$-th timestep and the $L'$-th block as follows:
\begin{gather}
    H_{k}(L', T') = \sum^{L'-1}_{l=1}\sum^{T'}_{t=1}h^{l, t}_{k}.
    \label{eq:M_score}
\end{gather}

\sys{} masks tokens with $H_{k}(L', T') \geq 1 - \epsilon$ in each block.  
If the 
$k$-th token is halted at the $L'$-th block and $T'$-th timestep, it remains zeroed out from the $L'+1$ block onward in the $T'$-th timestep.
Fig.~\ref{fig:main_example} illustrates a token-level merging and masking example of AT-SNN.

Based on the defined halting score, we propose a new loss function that allows \sys{} to determine the required number of tokens according to the input image during training. 
We define $\mathcal{N}^{t}_{k}$ as the index of the block where the $k$-th token halts at the $t$-th timestep, which is obtained by 
\begin{equation}
    \mathcal{N}_{k}^{t} = \underset{l \leq L}{\mathrm{arg\,min}}\ H_{k}(l, t) \geq 1 - \epsilon,
\end{equation}
where $\epsilon$ is a constant value that determines the threshold for the halting score. 

Additionally, we define an auxiliary variable, remainder, to track the remaining amount of halting score for each token until it halts at each timestep and layer as follows:
\begin{equation}
    r^{l,t}_{k} = 1 - H_{k}(l, t).
\end{equation}

Then, we define the halting probability of each token at each timestep and block as follows:
\begin{equation}
    p_{k}^{l, t} = \Bigg\{ \begin{array}{l}
        h_{k}^{l,t}\text{ if }t = \{1, ..., T\}\text{ and }l < \mathcal{N}^{t}\\
        r_{k}^{l,t}\text{ if }t = \{1, ..., T\}\text{ and }l = \mathcal{N}^{t}\\
        0\text{ otherwise }
    \end{array}
\end{equation}

According to the definitions of $h^{l,t}_{k}$ and $r^{l,t}_{k}$, $0 \leq p^{l, t}_{k} \leq 1$ holds. 

Based on the previously defined halting module and probability, we propose the following loss functions for training \sys{}. First, we apply a mean-field formulation (halting-probability weighted average of previous states) to the output at each block and timestep, accumulating the results. Therefore, the classification loss function $\mathcal{L}_{task}$ is defined as follows:
\begin{equation}
    \mathcal{L}_{task} = \mathcal{C}(FC(\frac{1}{TK}\sum^{T}_{t=1}\sum^{K}_{k=1}\sum^{L}_{l=1}\mathcal{T}^{l, t}_{k} \cdot p^{l, t}_{k})),
    \label{eq:cross_entrophy}
\end{equation}
where $\mathcal{C}$ denotes the cross-entropy loss. 

Next, we propose a loss function to encourage each token to halt at earlier timesteps and blocks, using fewer computations.
We defined the ponder loss $\mathcal{L}_{ponder}$ as follows:
\begin{equation}
    \mathcal{L}_{ponder} = \frac{1}{TK} \sum_{t=1}^{T}\sum^{K}_{k=1}(\mathcal{N}^{t}_{k} + r_{k}^{\mathcal{N}^{t}_{k}, t}).
    \label{eq:ponder_loss}
\end{equation}

$\mathcal{L}_{ponder}$ consists of the average number of blocks over which each token accumulates its halting score and the average remainder at each timestep.
\begin{equation}
    \mathcal{L}_{overall} = \mathcal{L}_{task} + \delta_{p} \mathcal{L}_{ponder},
    \label{eq:final_loss}
\end{equation}
where $\delta_{p}$ is a parameter that weights $\mathcal{L}_{ponder}$. 
\sys{} is trained to minimize $\mathcal{L}_{overall}$.

\subsection{Flexible Halting Threshold}
\label{subsec:AT2-SNN}

\sys{} adaptively determines the number of tokens to halt for each input image during training.
However, during inference, there remains a trade-off between the number of tokens to halt and accuracy.
To address this, we introduce \syss{}, a method that provides control-knob between the number of tokens to halt and accuracy by adjusting the halting threshold parameter $\epsilon$ during inference.
By increasing the value of $\epsilon$, \syss{} halts more tokens at earlier blocks or timesteps, leading to reduced energy consumption and accuracy.

\section{Experiments}
\label{sec:evaluation}

\begin{figure}[t!]
    \centering
    \includegraphics[width=1\linewidth]{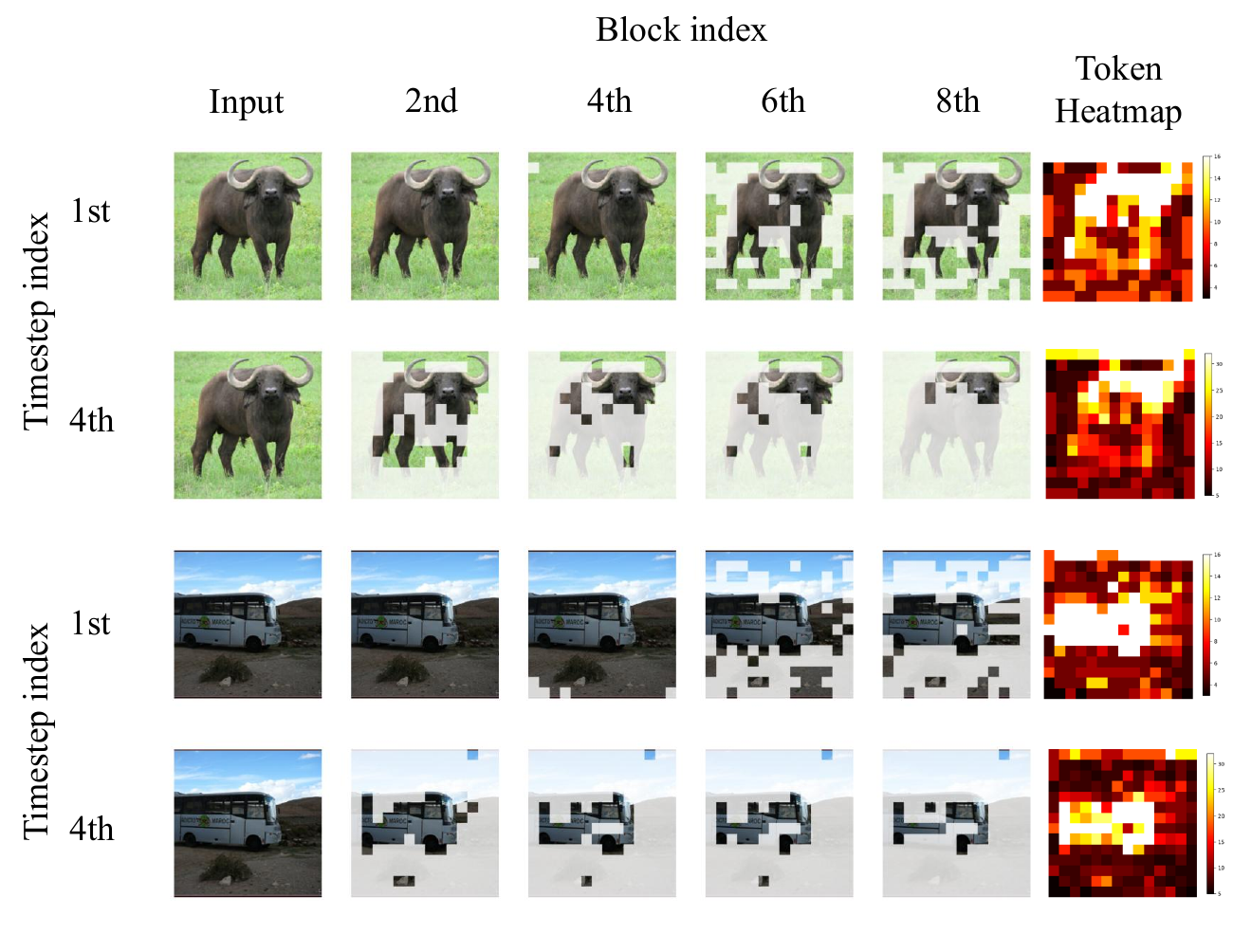}
    \caption{Example of halted tokens across different timesteps and blocks on \syss{} (based on Spikingformer-8-384) with ImageNet. Tokens that are halted with a shaded (non-white) overlay.}
    \label{fig:timesteps_blocks_sample}
\end{figure}
\begin{figure}[t!]
    \centering
    \includegraphics[width=1\linewidth]{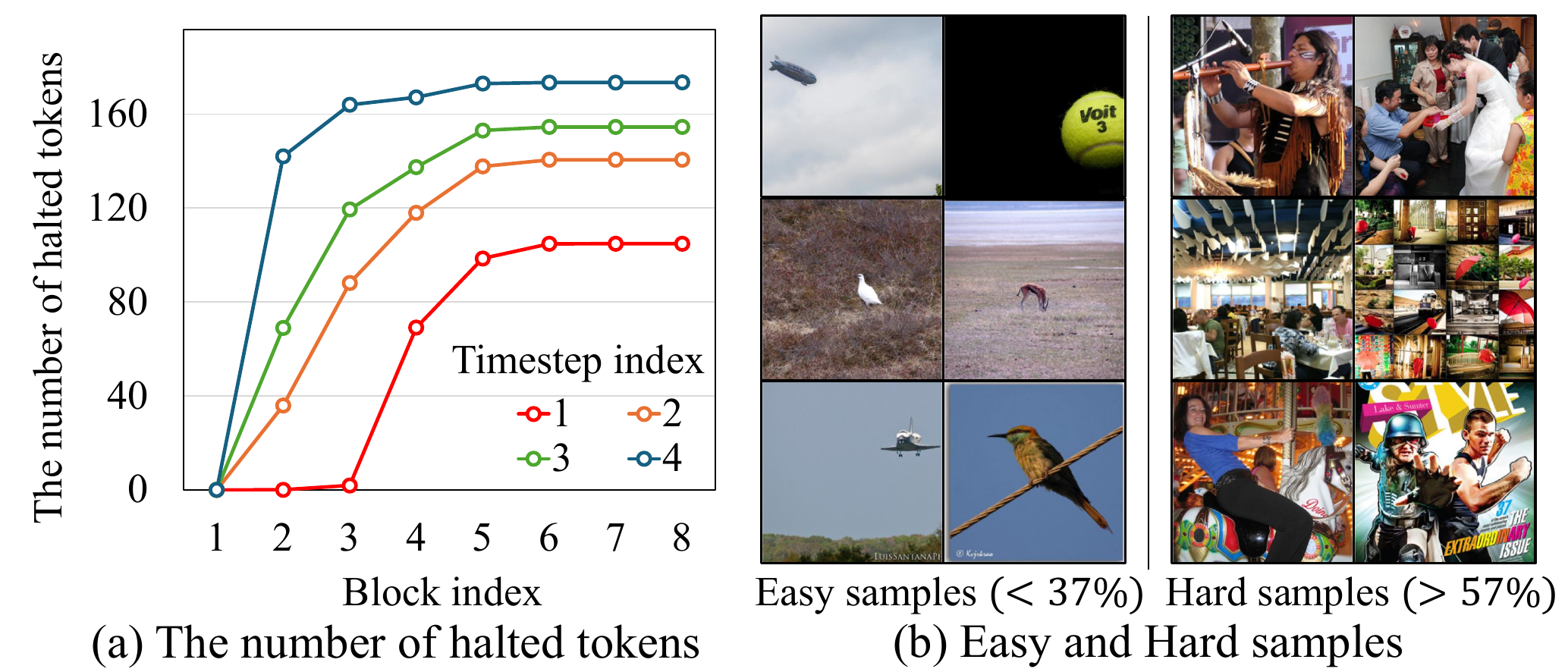}
    \caption{(a) The number of halted tokens across different blocks and timesteps, and (b) visual comparison of hard and easy samples in classification on \syss{} (based on Spikingformer-8-384) with ImageNet.}
    \label{fig:token_sample}
\end{figure}

We first analyze the qualitative and quantitative results to assess how efficiently \sys{} reduces tokens for the input images. 
Then, we conduct a comparative analysis to evaluate how effectively \sys{} reduces tokens in terms of accuracy, comparing it with existing methods, and analyze how the reduced tokens by \sys{} impact energy consumption. 
Finally, we discuss the properties required for \sys{}'s two-dimensional ACT to efficiently process tokens through an ablation study.

\begin{table*}[t]
\begin{center}
\caption{Main experiment results on ImageNet.}
\label{tab:main_exp}
\small
\begin{tblr}{
  colspec = {cccccc},
  row{23-26} = {gray!30},
  rowsep = 0pt,
}
\hline
Method                            & Architecture          & Param (M) & Timestep & Energy (mJ)           & Acc                \\ \hline
Hybrid training~\cite{RSP20}      & ResNet-34             & 21.79     & 250      & -                     & 61.48              \\
STBP-tdBN~\cite{ZWD21}            & ResNet-34             & 21.79     & 6        & 6.39                  & 63.72              \\
\SetCell[r=2]{m} TET~\cite{DLZ22} & Spiking-ResNet-34     & 21.79     & 6        & -                     & 64.79              \\
                                  & SEW ResNet-34         & 21.79     & 4        & -                     & 68.00              \\
\SetCell[r=2]{m} Spiking ResNet~\cite{HTP21} 
                                  & ResNet-34             & 21.79     & 350      & 59.30                 & 71.61              \\
                                  & ResNet-50             & 25.56     & 350      & 70.93                 & 72.75              \\
\SetCell[r=4]{m} SEW ResNet~\cite{FYC21} 
                                  & SEW ResNet-34         & 21.79     & 4        & 4.04                  & 67.04              \\
                                  & SEW ResNet-50         & 25.56     & 4        & 4.89                  & 67.78              \\
                                  & SEW ResNet-101        & 44.55     & 4        & 8.91                  & 68.76              \\
                                  & SEW ResNet-152        & 60.19     & 4        & 12.89                 & 69.26              \\
MS-ResNet~\cite{HWD21}            & ResNet-104            & 44.55+    & 5        & -                     & 74.21              \\
\SetCell[r=2]{m} Att MS ResNet~\cite{YZZ23} 
                                  & Att-MS-ResNet-18      & 11.87     & 1        & 0.48                  & 63.97              \\
                                  & Att-MS-ResNet-34      & 22.12     & 1        & 0.57                  & 69.15              \\ \hline
ANN                               & Transformer-8-512     & 29.68     & -        & 38.34                 & 80.80              \\ \hline
Spikformer~\cite{ZZH22}           & Spikformer-8-768      & 66.34     & 4        & 21.48                 & 74.81              \\ \hline
\SetCell[r=2]{m} OST~\cite{SSX24} & OST-8-384             & 19.36     & 1  & 4.63                  & 72.42              \\
                                  & OST-8-512             & 33.87     & 1  & 6.92                  & 74.97              \\ \hline
\SetCell[r=2]{m} Spikingformer~\cite{ZYZ23} 
                                  & Spikingformer-8-384   & 16.81     & 4        & \textbf{4.69}                  & \textbf{72.45}              \\
                                  & Spikingformer-8-512   & 29.68     & 4        & \textbf{7.46}                  & \textbf{74.79}              \\ \hline
\SetCell[r=2]{m} STATA~\cite{ZWY24}       
                                  & Spikingformer-8-384   & 16.82     & 4        & 4.33$^{*}$            & 67.65$^{*}$        \\ 
                                  & Spikformer-8-768      & -         & 4        & 11.16                 & 74.03              \\ \hline\hline
\SetCell[r=2]{m} \sys{}           & Spikingformer-8-384   & 16.81     & 4        & \textbf{3.81 (-18.8\%)}        & \textbf{73.45 (\textbf{$\uparrow$})} \\
                                  & Spikingformer-8-512   & 29.68     & 4        & \textbf{7.16 (-4.02\%)}        & \textbf{75.96 (\textbf{$\uparrow$})} \\
\SetCell[r=2]{m} \syss{}          & Spikingformer-8-384   & 16.83     & 4        & \textbf{3.28 (-30.1\%)}        & \textbf{72.61 (\textbf{$\uparrow$})} \\
                                  & Spikingformer-8-512   & 29.68     & 4        & \textbf{5.73 (-23.19\%)}       & \textbf{75.13 (\textbf{$\uparrow$})} \\ \hline
\end{tblr}
\end{center}
\end{table*}

\paragraph{Implementation details.}
We implement the simulation on Pytorch and SpikingJelly~\cite{FCD23}.
All experiments in this section are conducted on Spikformer~\cite{ZZH22} and Spikingformer~\cite{ZYZ23} on RTX NVIDIA A6000 GPUs.
Note that \sys{} is applicable to other SNN-based vision Transformers with direct training.
We first train each model by replacing its original SPS module with the proposed I-SPS, and use the resulting model as a pre-trained model for applying the proposed two-dimensional ACT.
Subsequently, we retrain the models using the loss function defined in Eq.~\eqref{eq:final_loss}.
We use automatic-mixed precision (AMP)~\cite{MNA17} for training acceleration and surrogate module learning (SML)~\cite{DLL23} method to mitigate the gradient errors of SNNs.
For a fair comparison, we trained several existing methods (e.g., Spikformer, Spikingformer, and STATA\footnote{As the official implementation is not publicly available, we re-implemented the method based on the descriptions in the original paper and made our best effort to reproduce it faithfully.}) on our computing environment, and these models are marked with an asterisk (*) in 
Tables~\ref{tab:main_exp} and~\ref{tab:cifar_exp}.
We evaluate our method for the classification task on CIFAR-10, CIFAR-100~\cite{KRH09}, and ImageNet~\cite{DDS09}.

\subsection{Analysis}
\label{subsec:eval_analysis}

\paragraph{Qualitative results.}
For visualization of \syss{}, we use Spikingformer-8-384 with eight blocks per timestep, trained on ImageNet. 
Each input image contains 196 tokens ($14\times14$).
Fig.~\ref{fig:timesteps_blocks_sample} visualizes how tokens are halted over timesteps and blocks. 
Since \sys{} accumulates halting scores in two dimensions (blocks and timesteps), more tokens are halted as the block and timestep indices increase.
With four timesteps and eight blocks, the maximum processed count for each token is 32, where brighter regions indicate more processing, and darker regions indicate less (i.e., halted earlier). 
Tokens from the less informative background are halted first, with an increasing number of tokens being halted over time.

\paragraph{Quantitative results and classification difficulty.}
Fig.~\ref{fig:token_sample}(a) shows the number of tokens halted per block and timestep. 
As visualized in Fig.~\ref{fig:timesteps_blocks_sample}, more tokens are halted as the block and timestep indices increase. 
Due to the two-dimensional halting policy of \syss{}, more tokens halt as the number of timesteps increases.
Figure~\ref{fig:token_sample}(b) visualizes samples correctly classified by \syss{}, comparing those that use more tokens versus those that use fewer tokens. 
On average, easy samples utilize 37\% or fewer of all tokens per block, while hard samples use 57\% or more of all tokens per block.
We observe that \syss{} uses fewer tokens when the object in the image is clearly separated from the background and other objects. 

\begin{table*}[t]
\begin{center}
\caption{Experiment results on CIFAR-10/CIFAR-100.}
\label{tab:cifar_exp}
\small
\begin{tblr}{
  colspec = {cccccc},
  rowsep = 0pt,
  row{12,16} = {gray!30},
}
\hline
Method                       & Architecture          & Param (M) & Timestep & Energy (mJ)                      & Acc                         \\ \hline
STBP-tdBN~\cite{ZWD21}       & ResNet-19             & 12.63     & 4        & -                                & 92.9/70.9                   \\
AutoSNN~\cite{NMP22}         & AutoSNN (C=128)       & 21        & 8        & -                                & 93.2/69.2                   \\
SpikeDHS$^{D}$~\cite{CLZ22}  & SpikeDHS-CLA (n3s1)   & 14        & 6        & -                                & 95.4/76.3                   \\
Hybrid training~\cite{RSP20} & VGG-11                & 9.27      & 125      & -                                & 92.2/67.9                   \\
Diet-SNN~\cite{RNR20}        & ResNet-20             & 0.27      & 10/5     & -                                & 92.5/64.1                   \\
TET~\cite{DLZ22}             & ResNet-19             & 12.63     & 4        & -                                & 94.4/74.5                   \\
ANN-to-SNN~\cite{DSG21}      & ResNet-20             & 10.91     & 32       & -                                & 93.3/68.4                   \\ \hline
ANN                          & Transformer-4-384     & 9.32      & -        & 4.25                             & 96.7/81.0                   \\ \hline
Spikformer~\cite{ZZH22}      & Spikformer-4-384      & 9.32      & 4        & \textbf{0.74$^{*}$/0.89$^{*}$}   & \textbf{94.8$^{*}$/77.3$^{*}$}     \\
STATA~\cite{ZWY24}           & Spikformer-4-384      & -         & 4        & -                                & 95.2/77.9                   \\ \hline
\syss{}                      & Spikformer-4-384      & 9.32      & 4        & \textbf{0.40/0.50}               & \textbf{95.2/77.9}                   \\ \hline
OST~\cite{SSX24}             & OST-4-384             & 11.37     & 1        & 0.46                             & 95.6/78.8                   \\ \hline
Spikingformer~\cite{ZYZ23}   & Spikingformer-4-384   & 9.32      & 4        & \textbf{0.42$^{*}$/0.50$^{*}$}   & \textbf{95.7$^{*}$/79.4$^{*}$}     \\
STATA~\cite{ZWY24}           & Spikingformer-4-384   & -         & 4        & -                                & 95.8/79.9                   \\ \hline
\syss{}                      & Spikingformer-4-384   & 9.32      & 4        & \textbf{0.37/0.46}               & \textbf{95.8/79.4}                   \\ \hline
\end{tblr}
\end{center}
\end{table*}

\subsection{Comparison to Prior Art}
\label{subsec:eval_comparison}

We evaluate STAS against SNN methods based on both CNNs (e.g., VGG, ResNet) and Transformers (e.g., Spikformer, Spikingformer). To benchmark against other dynamic computation techniques for SNN-based ViTs, we also compare our results with those of OST and STATA. We measured the energy consumption\footnote{Following the widely accepted measurement methods in previous SNN studies~\cite{ZZH22,ZYZ23}, the equation for calculating energy consumption is provided in the supplement.} and accuracy of each model during inference on ImageNet (in Table~\ref{tab:main_exp}) and CIFAR-10/CIFAR-100 (in Table~\ref{tab:cifar_exp}).

\paragraph{ImageNet}
We trained \sys{} on the Spikingformer-8-384 and Spikingformer-8-512 models. We set hyper-parameters as $\alpha=5$, $\beta=-25$, and $\delta_p=10^{-4}$. 
To compare against a static token-dropping method, we implemented STATA\footnote{Same as Footnote 1.} and evaluated its performance. 
As shown in Table~\ref{tab:main_exp}, Transformer-based methods generally outperform CNN-based ones. 
On the Spikingformer-8-384, STATA reduces some energy but incurs a significant accuracy drop because it drops a fixed ratio of tokens without considering timesteps. 
In contrast, \sys{} reduces energy consumption while achieving even higher accuracy than the original Spikingformer. Furthermore, by adjusting the halting threshold $\epsilon$, we can create a variant, \syss{}, which trades some accuracy for greater energy savings. 
When configured for significant energy savings, STAS reduces the energy consumption of the original Spikingformer by 18.8\% to 30.1\% while maintaining a comparable or even slightly higher accuracy.

\paragraph{CIFAR-10/CIFAR-100}
We trained \sys{} on Spikformer-4-384 and Spikingformer-4-384. We set hyper-parameters as $\alpha=-5, \beta=0, \delta_p=10^{-3}$ for Spikformer, and $\alpha=5, \beta=-25, \delta_p=10^{-3}$ for Spikingformer. For a fair comparison, we adjusted the halting threshold $\epsilon$ to create STAS variants tuned to the accuracy levels of the original models. For the Spikformer, we achieved substantial energy reductions of 45.9\% on CIFAR-10 and 43.8\% on CIFAR-100, respectively, while attaining higher accuracy. On the Spikingformer, STAS also achieved higher accuracy while reducing energy by 11.9\% on CIFAR-10 and 8.0\% on CIFAR-100.

\begin{table}[]
\small
\begin{center}
\caption{Ablation study on Spikformer-4-384 and Spikingformer-4-384 with CIFAR-100.}
\label{tab:ablation}
\resizebox{\columnwidth}{!}{
\begin{tabular}{cccc|cc}
\hline
Architecture & I-SPS       & $\epsilon$  & Accumulation               & Avg. tokens                 & Acc (\%)         \\ \hline
\multirow{4}{*}{\small Spikformer}
             & \ding{55}   & \ding{55}   & \circled{B}                & $\times$0.60                & 78.0             \\
             & \ding{55}   & \ding{55}   & \circled{T} + \circled{B}  & $\times$0.63                & 77.3             \\
             & \ding{51}   & \ding{55}   & \circled{T} + \circled{B}  & \textbf{$\times$0.46}       & \textbf{78.1}    \\ 
             & \ding{51}   & \ding{51}   & \circled{T} + \circled{B}  & \textbf{$\times$0.42}       & \textbf{77.9}    \\ \hline
\multirow{4}{*}{\small Spikingformer}
             & \ding{55}   & \ding{55}   & \circled{B}                & $\times$0.65                & 78.5             \\
             & \ding{55}   & \ding{55}   & \circled{T} + \circled{B}  & $\times$0.95                & 77.4             \\
             & \ding{51}   & \ding{55}   & \circled{T} + \circled{B}  & \textbf{$\times$0.70}       & \textbf{79.9}    \\ 
             & \ding{51}   & \ding{51}   & \circled{T} + \circled{B}  & \textbf{$\times$0.50}       & \textbf{78.5}    \\ \hline
\end{tabular}
}
\end{center}
\end{table}

\subsection{Ablation Studies}
\label{subsec:eval_ablation}

We evaluate the impact of I-SPS and the accumulation methods on the accuracy and energy efficiency of \sys{}.
Table~\ref{tab:ablation} shows the average number of tokens used per block and the corresponding accuracy with and without each component.
All experiments are conducted on the Spikformer-4-384 model using the CIFAR-100.

\paragraph{I-SPS vs SPS.}
Table~\ref{tab:ablation} presents the token usage and accuracy of \sys{} with and without I-SPS.
With I-SPS, \sys{} achieves higher accuracy (77.3\% vs. 78.1\%) while using fewer tokens ($\times$0.63 vs. $\times$0.46).
This improvement arises because, as shown in Fig.~\ref{fig:SPS}(c), I-SPS encourages similarity among inputs across consecutive timesteps, enabling more efficient application of ACT.

\paragraph{Two- vs one-dimensional halting.}
Table~\ref{tab:ablation} compares the halting score accumulation methods on CIFAR-100: one that accumulates only across one dimension (\circled{B}, block-level only) and another that accumulates scores across two dimensions (\circled{T} + \circled{B}, both timestep and block-levels as per Eq.~\eqref{eq:M_score}). 
As shown in Table~\ref{tab:ablation}, the two-dimensional halting mechanism achieves higher accuracy (78.0\% vs 78.1\%) while removing more tokens ($\times$0.60 vs $\times$0.46) compared to the one-dimensional halting. 
This is because, by definition, the LHS of Eq.~\eqref{eq:M_score} becomes larger under two-dimensional halting than under one-dimensional halting, which in turn increases the LHS of Eq.~\eqref{eq:ponder_loss}, leading to more tokens being halted. 
Furthermore, the \syss{} variant maximizes this halting effect, achieving even greater token reduction ($\times$0.42 and $\times$0.50 for Spikformer and Spikingformer, respectively).

\section{Conclusion}
\label{sec:conclusion}

In this paper, we addressed the fundamental two-dimensional (spatio-temporal) adaptive computation challenge inherent to SNN-based ViTs. 
We first identified that the efficacy of dynamic halting is fundamentally constrained by the static architecture's lack of temporal similarity. 
To resolve this, we proposed STAS, a framework that co-designs a static architectural module (I-SPS) with a dynamic halting policy (A-SSA) to enable accuracy-aware token halting across both spatial and temporal axes. 
Our experiments on CIFAR-10, CIFAR-100, and ImageNet demonstrate the effectiveness of this synergistic approach: STAS significantly improves the accuracy-energy trade-off, reducing energy consumption by up to 45.9\%, 43.8\%, and 30.1\%, respectively, while simultaneously enhancing accuracy.

\bibliography{aaai2026}

\end{document}